\documentclass[letterpaper]{article} 
\usepackage{aaai24}  
\usepackage{times}  
\usepackage{helvet}  
\usepackage{courier}  
\usepackage[hyphens]{url}  
\usepackage{graphicx} 
\usepackage{natbib}  
\usepackage{caption} 
\usepackage{algorithm}
\usepackage{algorithmic}
\usepackage{amsmath}
\usepackage{mathtools}
\usepackage{amsthm}
\usepackage{amsfonts}
\usepackage{caption}
\usepackage{multirow}
\usepackage{subfigure}
\usepackage{xcolor}
\usepackage{comment}
  
 \newcommand{\cD}{\mathcal{D}}

 \newcommand{\cL}{\mathcal{L}}

\newcommand{\cM}{\mathcal{M}}

\DeclareMathOperator*{\argmin}{\textit{argmin }}

\usepackage{newfloat}
\usepackage{listings}
\DeclareCaptionStyle{ruled}{labelfont=normalfont,labelsep=colon,strut=off} 
\floatstyle{ruled}
\newfloat{listing}{tb}{lst}{}
\floatname{listing}{Listing}
\title{VeriCompress: A Tool to Streamline the Synthesis of Verified Robust Compressed Neural Networks from Scratch}
\author{
    Sawinder Kaur\textsuperscript{\rm 1}
    Yi Xiao\textsuperscript{\rm 1}
    Asif Salekin\textsuperscript{\rm 1}
}
\affiliations{
    \textsuperscript{\rm 1}Syracuse University, USA \\
    sakaur@syr.edu, yxiao54@syr.edu, asalekin@syr.edu
}

\usepackage{bibentry}

\begin{document}

\maketitle
\begin{abstract}

AI's widespread integration has led to neural networks (NNs) deployment on edge and similar limited-resource platforms for safety-critical scenarios. Yet, NN's fragility raises concerns about reliable inference. Moreover, constrained platforms demand compact networks. This study introduces \textcolor{violet}{VeriCompress}, a tool that automates the search and training of compressed models with robustness guarantees. These models are well-suited for safety-critical applications and adhere to predefined architecture and size limitations, making them deployable on resource-restricted platforms. The method trains models $2-3$ times faster than the state-of-the-art approaches, surpassing relevant baseline approaches by average accuracy and robustness gains of $15.1$ and $9.8$ percentage points, respectively. When deployed on a resource-restricted generic platform, these models require $5-8$ times less memory and $2-4$ times less inference time than models used in verified robustness literature. Our comprehensive evaluation across various model architectures and datasets, including MNIST, CIFAR, SVHN, and a relevant pedestrian detection dataset, showcases \textcolor{violet}{VeriCompress}'s capacity to identify compressed verified robust models with reduced computation overhead compared to current standards. This underscores its potential as a valuable tool for end users, such as developers of safety-critical applications on edge or Internet of Things platforms, empowering them to create suitable models for safety-critical, resource-constrained platforms in their respective domains.

\end{abstract}

\section{Introduction}
\label{sec:introduction}


Neural Networks (NNs) are vulnerable to small deviations in the input, i.e., imperceptible adversarial perturbation defined by the perturbation budget $\varepsilon$, which can lead to inaccurate inferences~\cite{goodfellow2015explaining}. Such vulnerability may have catastrophic effects \cite{make2040031} in safety-critical real-world applications such as medical diagnosis or autonomous driving, where the input data may often be slightly changed due to sensor noise, measurement errors, or adversarial attacks.

\par 
Although \citet{madry2018towards} introduced PGD adversarial training to boost robustness, \citet{Gowal2018OnTE} revealed that such training doesn't guarantee robustness. Adversarial samples were discovered near the supposedly robust ones, raising concerns for safety-critical applications such as ACAS-Xu, pedestrian detection, and healthcare \cite{reluplex,robust-pedestrain-detection,health-care}.

To address this, \textit{Verified robustness} training methods~\cite{reluplex,Gowal2018OnTE,crown-ibp,auto-lirpa} were developed, providing robustness guarantees. These methods certify the absence of adversarial samples near robust ones. However, they achieve verified robustness at the expense of accuracy \cite{crown-ibp,auto-lirpa}. Despite this trade-off, ensuring robustness within specific feature space regions is crucial for the secure implementation of sensitive applications. E.g., these robustness assurances have been used to partition the feature space into regions of varying robustness for ACAS-Xu \cite{julian2019verifying}. This division facilitates training hierarchical models that exhibit high robustness in different feature space regions, thereby enhancing the reliability of neural network decisions in safety-critical contexts. Detailed background on verified robustness training is in Appendix A.

\par
A challenge that yet remains unresolved is that many real-world safety-critical applications demand NNs to be deployed on resource-constraint platforms, such as drones, self-driving cars, smart-watches, etc., which kindles the need for highly compressed models, requiring significantly low computation and storage resources, all while achieving accuracy and verified robustness comparable to the denser counterparts developed in state-of-the-art. Finding such compressed verified robust NN architectures requires extensive computations, compounded by model size limitations. Moreover, the architecture varies based on applications, necessitating domain knowledge. E.g., image classification leans on convolutional layers (CNNs), while image captioning benefits from sequential architectures (RNNs).

\par 
This work aims to automate the exploration of compressed networks while concurrently training a verified robust model within the confines of a \emph{parameter-budget}—an estimated count of parameters for the required compressed model suitable for resource-constrained device deployment. The paper presents \textcolor{violet}{VeriCompress}, a tool enabling users to choose a desired architecture type in the form of an existing complex/dense randomly initialized model, defined as backbone architecture, and a parameter budget. \textcolor{violet}{VeriCompress} then automatically identifies a compressed architecture, i.e., a sub-network within the specified backbone that meets the parameter budget while attaining similar accuracy and verified robustness compared to denser models.

\par
\textit{Research Gap being Addressed:} 
Amid various pruning techniques, structured model pruning stands out for effectively decreasing model size and computational resources \cite{no-time-reduction} (refer to Appendix B for details). However, the leading pruning methods, Hydra \cite{sehwag2020hydra} and Fashapley \cite{kang2023fashapley}, designed to ensure verified robustness, exhibit suboptimal performance when applied to structured pruning. Additionally, these techniques necessitate prior training of the original dense model, leading to substantial resource and time burdens. In contrast, \textcolor{violet}{VeriCompress} streamlines the process by initiating from a randomly initialized large model (backbone), considerably reducing training time and resource overhead. Moreover, \textcolor{violet}{VeriCompress} simultaneously achieves relatively higher \emph{generalizability} compared to state-of-the-art robust pruning approaches \cite{sehwag2020hydra,kang2023fashapley}. In this work, we refer to \emph{model generalizability} as the ability of the model to project high-standard accuracy and verified robustness simultaneously.

\par 
The paper’s contributions are summarized below:
\begin{enumerate}
    \item This is the first work to demonstrate that verified robust compressed neural networks can be trained from scratch while outperforming baseline works \cite{sehwag2020hydra,kang2023fashapley} by an average of $15.1$ and $9.8$ percent points in accuracy and verified robustness.

    \item \textcolor{violet}{VeriCompress} extracts compressed NNs taking $2-3$ times less training time than state-of-the-art structured pruning approaches in the same platform. 


    \item The compressed models thus achieved, when deployed on a resource-constrained generic platform, such as Google Pixel 6, require about $5-8$ times less memory and $2-4$ times less inference time compared to the starting architecture.

    \item Our empirical study demonstrates that \textcolor{violet}{VeriCompress}  performs effectively on benchmark datasets (CIFAR-10, MNIST, \& SVHN) and application-relevant \emph{Pedestrian Detection} \cite{pedestrian-dataset} dataset, and model architecture combinations used in the literature.
    
\end{enumerate}

The following sections provide related work discussion, details of the approach, and supporting evaluations. Additionally, Appendices include relevant background.

\section{Related work}

With the recent advances in verified robustness, to the best of our knowledge, two recent works have aimed at achieving compressed networks that exhibit verified robutness~\cite{sehwag2020hydra,kang2023fashapley}. 

\par 
\citet{sehwag2020hydra} (Hydra) proposed a three-step robustness-aware model training approach accounting for both adversarial robustness and verified robustness (also known as certified robustness). The pre-train step aims to generate a robust dense model, which then undergoes prune-training to learn a robustness-aware importance score for each parameter, which guides the final pruning mask. The finetuning step updates the retained weights to recover the model's performance. Hydra sets an equal pruning ratio for all the layers to obtain the pruning mask, which ignores the fact that different layers hold different importance toward model inference. 
\par 
\citet{kang2023fashapley} developed FaShapley following Hydra's framework with the difference in the prune-training step. The update to a parameter's importance scores is equivalent to the magnitude of the product of weight and gradient, which is an efficient approximation for its shapely value. The final removal of weights can be done in a structured or unstructured manner based on the importance scores (Shapley values) considered globally.
\par

This paper considers Hydra~\cite{sehwag2020hydra} and Fashapley~\cite{kang2023fashapley} as baselines and shows their performance comparisons with \textcolor{violet}{VeriCompress}.

\section{Parameter-Budget-Aware Architecture Search for Verified Robust Models}\label{presented-approach}

\textcolor{violet}{VeriCompress} is a tool that takes an initially over-parameterized randomly initialized NN architecture, serving as the backbone (essentially representing the desired architecture type), along with a parameter budget, and it then searches and identifies a verified robust model within the confinement of the backbone and specified size. The approach adapts the dynamic sparse training paradigm \cite{nest,itop}, which allows the model to learn sparse networks from scratch (discussed in Appendix B).


\subsection{Problem Definition - The Learning Objective} \label{problem-definition}
For a given randomly initialized backbone architecture $\cM_{\theta}$ with $k$ parameters and a parameter budget $k'$, the objective is to train a verified robust sparse network $\cM_{\theta^\downarrow}$ which uses only a subset of available parameters (of the backbone), that is,   $|\theta^\downarrow|_0 = k'$ and $k'<<k$. The compressed model thus achieved should be comparable in generalizability to the backbone model being trained considering all $k$ parameters, the generalizability being measured in terms of accuracy and verified robustness at the perturbation budget $\varepsilon$. 

\par 
\subsection{Detailed Discussion of the \textcolor{violet}{VeriCompress} Approach}\label{approach}

\par 
In order to achieve the desired effect, during the compressed network search, a model element (kernels and nodes) can either be in an \emph{active} or \emph{dormant} state. All the parameters associated with a dormant element are set to zero. However, the presence of even a single non-zero parameter makes the element active. It is important to note that only the active elements contribute towards the inference during the forward pass. However, during the backward pass, the gradient is computed for both active and dormant elements, which allows the model to explore new structures when model weights are updated. 

\begin{algorithm}[ht]
\begin{algorithmic}[1]
\small{
  \caption{\textcolor{violet}{VeriCompress}}
  \label{alg:learning}
  }
  \REQUIRE{$\cM_{\theta}$: Backbone architecture
  $~\diamond~ k'$: Target parameter budget
  $~\diamond~ \cD=(X_i, y_i)_{i=1}^m$: Dataset
     $~\diamond~  \varepsilon_\text{max}$: Maximum Perturbation
     $~\diamond~  T$: Number of training epochs
     $~\diamond~  (s,l)$: Start and length of perturbation scheduler
     $~\diamond~$ \emph{T-exp}: Number of epochs to be used for expansion
     $~\diamond~$ \emph{seed}: A seed value}

     \STATE Randomly initialize the backbone architecture using \emph{seed}.
  \STATE $\cM_{\theta'} =\cM_{\theta^\downarrow} =$ Deactivate$(\cM_{\theta},k')$ 
  \FOR{epoch $t = [ 1,2, \ldots, T]$} 
    \STATE $\varepsilon_t =  \varepsilon$-Scheduler$(\varepsilon_\text{max}, t,s,l$)
      \FOR {minibatches $d \in \cD$} 
        \STATE $\cM_{\theta'} = \underset{\theta}\argmin  \cL_\text{train}(\cM_{\theta'},d,\varepsilon_t)$ 
    
    \ENDFOR
    \IF {$t\%$ \emph{T-exp} $== 0$} 
    \STATE $\cM_{\theta'} = \cM_{\theta^\downarrow} =$ Deactivate$(\cM_{\theta'},k')$  
    \ENDIF
  \ENDFOR
  \STATE Remove the dormant parameters from $\cM_{\theta^\downarrow} $
\vskip -2ex
\end{algorithmic}

\end{algorithm}

Algorithm \ref{alg:learning} describes the \textcolor{violet}{VeriCompress} procedure, that 
requires a backbone architecture $M_\theta$; a target parameter-budget $k'$; the dataset $\cD$; and the maximum perturbation $\varepsilon_\text{max}$. It also requires hyperparameters that vary for different datasets and include the number of training epochs $T$, the inputs for the $\varepsilon$-scheduler: $s$ and $l$, and the number of epochs used for exploration: \emph{T-exp}.


\par

The approach initiates by activating a random sub-network within the input backbone architecture with the specified parameter budget $k'$ (lines 1-2). 
This model undergoes training for $T$ epochs to minimize the loss $\cL_{\text{train}}$ (Equation \ref{eq:robust_loss_main}), aiming for verified robustness (lines 3-11). Each model update step allows the exploration of new parameters in the backbone, as even the dormant elements are also involved, resulting in an auxiliary model $\cM_{\theta'}$ (lines 5-7). 
Since there is no restriction on the capacity of the auxiliary network $\cM_{\theta'}$, it may comprise more than $k'$ parameters. To restrict the sub-network size to $k'$ parameters, the least important elements are deactivated every \emph{T-exp} ($<< T$) epochs (lines 8-10), resulting in compressed model $\cM_{\theta^\downarrow}$. 
The deactivation process involves evaluating the importance of each model element, as elaborated on in subsequent sections. 


This iterative process of activating and deactivating model elements enables the concurrent exploration and training of a subnetwork while adhering to the underlying backbone architecture and model sizes’ limitations \cite{itop,DST_reason}. Algorithm \ref{alg:learning} directs the training process by optimizing the training loss outlined in Equation \ref{eq:robust_loss_main}, ultimately producing a verified robust subnetwork.

Finally, after $T$ epochs, a robust sub-network $\cM_{\theta^\downarrow}$ with parameter budget $k'$ is identified, but the backbone architecture still contains dormant elements. For efficient deployment, the dormant elements are removed from the model resulting in a compressed model that demands fewer memory and computation time resources while exhibiting high verified robustness (line 12) \cite{torch-pruning}. \emph{Notably, this step does not require any further training.}



Different components and design choices of the \textcolor{violet}{VeriCompress} approach are discussed below:

\subsection{Training Loss}
\textcolor{violet}{VeriCompress} learns the model parameters to reduce the training loss as defined in state-of-the-art verified robust training approaches ~\cite{crown-ibp}:
\begin{equation}
    \cL_\text{train}(\cM_\theta,\cD,\varepsilon) = \sum_{(x_0,y) \in \cD} \max_{x\in\mathbb{B}(x_0,\varepsilon)}\cL(\cM_\theta(x),y),
    \label{eq:robust_loss_main}
\end{equation}
where $\cD$ is the dataset and $(x_0,y) \in \cD$. Here, $\mathbb{B}(x_0,\varepsilon)$ defines the $\ell_\infty$-ball of radius $\varepsilon$ around sample $x_0$. Intuitively, $\cL_\text{train}$ is the sum of maximum cross-entry loss in the neighborhood of each sample in $\cD$ for perturbation amount $\varepsilon$.  A detailed discussion about $\cL_\text{train}$ is provided in Appendix A.

The parameter weights, including the \emph{dormant} ones, are updated according to their corresponding gradients. 
The perturbation $\varepsilon$ used for computing $\cL_\text{train}$ is computed using the perturbation scheduler $\varepsilon-scheduler(\varepsilon_\text{max},t,s,l)$ (in line $4$ of Algorithm \ref{alg:learning}) as discussed below.

\paragraph{$\bullet$ Perturbation Scheduler $(\varepsilon$-scheduler ($\varepsilon_\text{max},t,s,l)$:}
Following state-of-the-art robust training approaches \cite{Gowal2018OnTE,crown-ibp,auto-lirpa}, \textcolor{violet}{VeriCompress} uses $\varepsilon-$scheduler to gradually increase perturbation starting at epoch $s$ for $l$ epochs. The gradual increase of epsilon prevents the problem of intermediate bound explosion while training; hence, it deems $\varepsilon-$scheduler necessary for effective learning (detailed in Appendix A).


\paragraph{$\bullet$ CROWN-IBP as the Sparse Regularizer in $\cL_\text{train}$}
\label{motivation-crown-ibp}
\citet{crown-ibp} noted that the verified robustness training mechanism proposed by \citet{wongNKolter} and \citet{scalable_fv} induce implicit regularization. CROWN-IBP~\cite{crown-ibp} incurs less regularization and shows an increasing trend in the magnitude of network parameters while training. However, according to our preliminary analysis, the implicit regularization caused by CROWN-IBP penalizes the network’s parameters, making them smaller compared to naturally trained networks (aiming for only accuracy), causing a high fraction of the parameters to be close to zero. Removal of such less significant parameters has minimal impact on model \emph{generalizability}.
\begin{figure}[t]
    \centering
    \subfigure[4-layer CNN (SVHN)]{\includegraphics[width=0.45\linewidth]{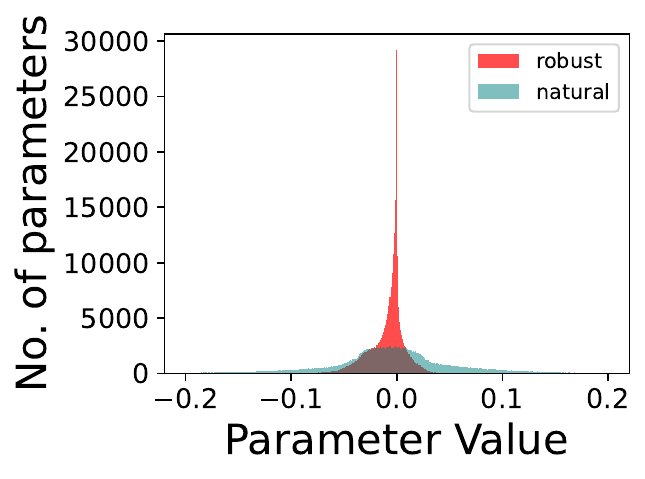}}\subfigure[4-layer CNN (CIFAR-10)]{\includegraphics[width=0.45\linewidth]{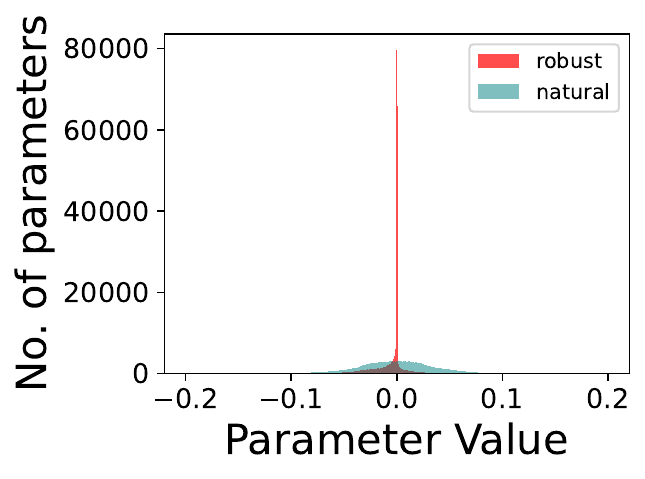}} 
    \subfigure[7-layer CNN (SVHN)]{\includegraphics[width=0.45\linewidth]{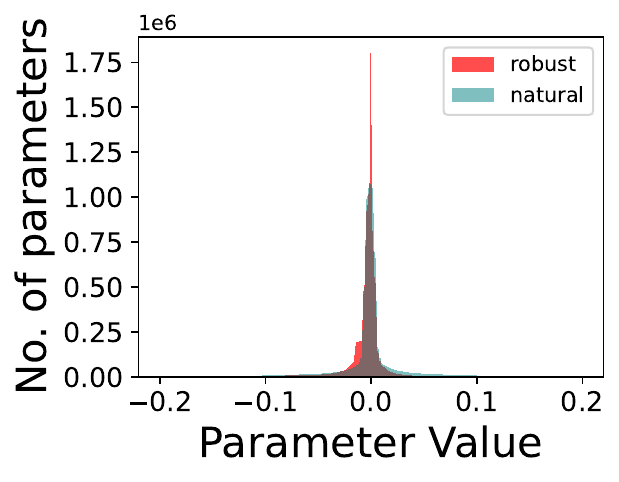}}
    \subfigure[7-layer CNN (CIFAR-10)]{\includegraphics[width=0.45\linewidth]{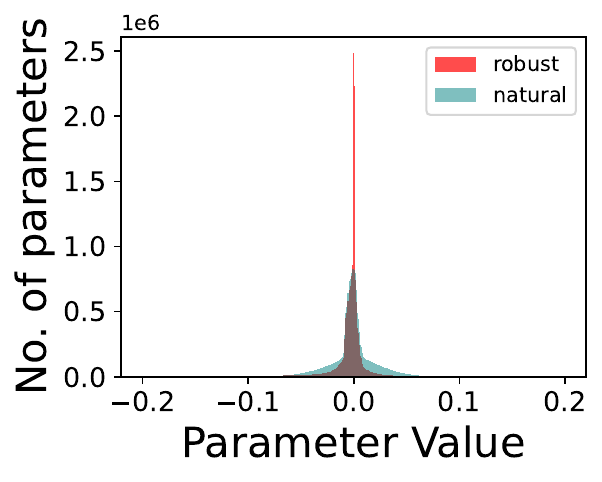}}
    \caption{Weight distributions for (a) 4-layer CNN (SVHN)  (b) 4-layer CNN (CIFAR-10) (b) 7-layer CNN (SVHN) (d) 7-layer CNN (CIFAR-10). The perturbation amount used for robust training is 2/255.}
    \vskip -3ex \label{fig:weight_distribution_cnn}
\end{figure}
\par 
For example, figure \ref{fig:weight_distribution_cnn} shows the weight distribution of networks trained to minimize verified robust (CROWN-IBP~\cite{crown-ibp}) and natural losses. As evident from the distribution of weights, a higher fraction of weights in verified robust models have very low magnitudes ($\approx 10^{-40}$). 
\par 
This observation suggests that minimizing the loss $\cL_\text{train}$ bounding by CROWN-IBP penalizes the network parameter magnitudes.
Existing sparse training approaches \cite{sparse-train-regularisation,sparse-train-l1} used regularization terms in their loss function to promote sparsity. Since the $\cL_\text{train}$ implicitly incorporates regularization, \textcolor{violet}{VeriCompress} does not include any additional regularization term.

\subsection{Parameter Deactivation based on Element Importance}
This step considers the importance of a node/kernel as a whole as the deciding factor for its state. All the parameters associated with the least significant nodes/kernels are set to zero.  \textcolor{violet}{VeriCompress} uses $\ell_2$-norm (following \cite{he2018soft}) of all the parameter weights corresponding to a kernel or node to decide the indices of kernels/nodes to be deactivated from a layer. Since the kernels/nodes belonging to different layers have different numbers of parameters associated with them, the norm of these kernels/nodes are not comparable globally~\cite{no_global_structured_pruning}.
Therefore, the deactivation process 
is applied layer-wise.

Thus, if $m_i$ represents the $i$th kernel/node of layer $l$ and $m_{i,j}$ represents $j^{th}$ parameter associated with it, element deactivation can be formulated as:
$\theta_l^\downarrow = [m_i| m_i\in p_l^{th} \text{ percentile of } \sqrt{\sum_j m_{i,j}^2}],$ and $\theta_l - \theta_l^\downarrow = 0,$ 
where $\theta_l^\downarrow$ represents active kernel/node retained at layer $l$ after deactivation and $p_l$ is the percentage of parameters to be deactivated at layer $l$. The value of $p_l$ is  decided based on Erd\H{o}s-R\'enyi-Kernel scaling  (used in \citep{evci2021rigging,raihan2020sparse-swat}).


\section{Experiments}
\label{sec:experiments}

The following section's evaluations demonstrate \textbf{(1)} The effectiveness of \textcolor{violet}{VeriCompress} in achieving verified robust models which outperform the state-of-the-art structured pruning approaches; \textbf{(2)} The scalability and generalizability of \textcolor{violet}{VeriCompress} with \emph{backbone network} variations, i.e., with different network complexities and architectures, such as Resnet, ResNext, and DenseNet; \textbf{(3)} The applicability of \textcolor{violet}{VeriCompress} to a practical application domain; and \textbf{(4)} the reduction in memory, training-time, and inference-time. 

\par 
\textbf{Metrics for evaluations:}\label{tab:eval-metrics} This paper utilizes the metrics used in the previous works for formal robustness verification ~\cite{sehwag2020hydra,kang2023fashapley} to evaluate the compressed models. The metrics are:
\begin{enumerate}
\item[$\bullet$] \emph{Standard Accuracy: } Percentage of benign samples classified correctly.
\item[$\bullet$] \emph{Verified Accuracy: } Percentage of benign samples which are certified to be robust using verified robustness mechanism IBP \cite{Gowal2018OnTE}, thus, measures the verified robustness of the model. Details are in Appendix A.
\end{enumerate}

\par 
\textbf{Datasets and Models:} For a fair comparison with state-of-the-art~\cite{kang2023fashapley}, benchmark dataset CIFAR-10~\cite{cifar10} and SVHN~\cite{svhn} are used. To demonstrate the effectiveness and generalizability of \textcolor{violet}{VeriCompress}, evaluations are shown for two more datasets: MNIST~\cite{lecun-mnisthandwrittendigit-2010} and Pedestrian Detection~\cite{pedestrian-dataset}. To demonstrate the versatility of the approach across different architectures, we evaluate (a) Two CNNs with varying capacity: 4-layer CNN and 7-layer CNN ~\cite{Gowal2018OnTE,crown-ibp}; (b) Network with skip connections: A 13-layer ResNet~\cite{scalable_fv}; (c) Two DNNs: DenseNet and ResNext ~\cite{auto-lirpa}. Notably, the models used in this work are used in state-of-the-art robust training approaches, as cited next to their names.

\textbf{Perturbation Amount:} For image classification datasets, the perturbations are introduced in $\ell_p-$ball of radius $\varepsilon$ (perturbation amount), $\mathbb{B}_p(x_0,\varepsilon)$ (defined in Section \ref{Background-and-motivation}). For a fair comparison, while comparing our results with baselines~\cite{sehwag2020hydra,kang2023fashapley}, we use the same perturbation budget used in these works: 2/255 for CIFAR-10 and SVHN. For MNIST, we use a $\varepsilon = 0.4$, the highest perturbation amount for MNIST (i.e., most difficult scenarios) used in the literature addressing formal verification and training ~\cite{Gowal2018OnTE,crown-ibp,scalable_fv,auto-lirpa}. For the Pedestrian Detection dataset, we use 2/255 as the perturbation amount.


\textbf{Hyperparameters:} \textcolor{violet}{VeriCompress} requires a set of hyperparameters as inputs: $\varepsilon-scheduler$ inputs $(T,s,l)$ and exploration length \emph{T-exp}. 
$T$ is the total number of training epochs, $s$ is the epoch number at which the perturbation scheduler should start to increment the amount of perturbation, and $l$ specifies the length of the schedule, that is, the number of epochs in which the perturbation scheduler has to reach the maximum amount of perturbation $\varepsilon_\text{max}$. 

\par These values are as follows: (1) For SVHN and CIFAR-10, we use $((T,s,l) = (330,15,150)$ (2) For MNIST, we use $((T,s,l) = (100,10,60)$, and (3) For Pedestrian Detection, the hyperparameters $(T,s,l) = (100,20,60)$ resulted in the models displaying the least errors empirically. 
Notably, using a fixed perturbation amount throughout training results in a trivial model.

To obtain the optimum values for the length of exploration \emph{T-exp}, Optuna~\cite{akiba2019optuna}, a parameter-tuning tool, is used in order to attain the combined objective of maximizing \emph{Standard} and \emph{Verified} accuracy. 

\textbf{Learning-Rate Decay:} \textcolor{violet}{VeriCompress} employs learning rate decay once the perturbation scheduler achieves the maximum perturbation. An initial high learning rate allows the dormant elements a fair chance to be considered toward the sub-network structure during the model exploration. 

\par 

\textbf{Formal Verification Mechanisms to Compute $\cL_\text{train}$ and \emph{Verified Accuracy}:} \textcolor{violet}{VeriCompress} training employs CROWN-IBP ~\cite{crown-ibp}. However, to be consistent with the state-of-the-art ~\cite{sehwag2020hydra,kang2023fashapley}, this section's presented evaluations use IBP to compute \emph{Verified accuracy}.

\subsection{Establishing the Effectiveness of \textcolor{violet}{VeriCompress}: Comparison with Baselines}
\label{Sec:comparisn_with_SOA}

This section compares the compressed models obtained via \textcolor{violet}{VeriCompress} with the state-of-the-art approaches: FaShapley~\cite{kang2023fashapley} (best-baseline) and Hydra~\cite{sehwag2020hydra}. Unfortunately, we could not reproduce the results for FaShapley with the provided code. Therefore, for a fair comparison, this section's evaluations are done with the same combination of model, dataset, and sparsity amounts as used in FaShalpley~\cite{kang2023fashapley} and compared with the baseline results as provided in the paper.

Table \ref{tab:structured-2-255} shows the comparison of \textcolor{violet}{VeriCompress} with Fashapley and Hydra for two architectures: 4-layer CNN and 7-layer CNN trained for CIFAR-10 and SVHN. The perturbation amount used for training and testing these models is $2/255$. The model size is shown in terms of the number of parameters where M stands for million. The percentage next to the model size represents the compressed model's size with respect to the backbone architecture. Across all model, dataset, and compression combinations, \textcolor{violet}{VeriCompress} attains an average increment of $15.1$ and $9.8$ percent points in \emph{Standard} and \emph{Verified} accuracy, respectively. The results shown in Table \ref{tab:structured-2-255} are an average of computations for three seed values followed by their error bars.

\begin{table}[t]
    \centering
    \resizebox{0.99\linewidth}{!}{
    \begin{tabular}{c|c|c|ll|ll}
        \hline
         &Model&Method & Standard & Verified& Standard & Verified \\
         \hline
         \multirow{10}{*}{\rotatebox[origin=c]{90}{CIFAR-10}}&\textbf{4-layer CNN}&Parameters& \multicolumn{2}{c|}{0.106 M (50\%)} & \multicolumn{2}{c}{0.085 M (40\%)} \\
         \cline{3-7}
         &Size=0.215 M &Hydra&44.2&32.5&32.5&22.0\\
         &Standard/&FaShapley&46.7&36.9&32.1&26.8\\
         &Verified &VeriCompress&\textbf{52.1} \small{($\pm$0.31)}&\textbf{42.2} \small{($\pm$0.15)} &\textbf{50.6} \small{($\pm$1.86)}&\textbf{41.0} \small{($\pm$1.93)}\\
         \cline{3-7}
         &54.3/42.9&$\Delta$&+5.4&+5.3&+18.5&+14.2\\
         \cline{2-7}
         &\textbf{7-layer CNN}&Parameters& \multicolumn{2}{c|}{3.295 M (20\%)} & \multicolumn{2}{c}{1.598 M (10\%)}\\
         \cline{3-7}
         &Size=17.190 M &Hydra&52.1&39.8&10.0&10.0\\
         &Standard/&FaShapley&55.5&43.9&19.5&14.2\\
         &Verified &VeriCompress&\textbf{59.1} \small{($\pm$0.55)}&\textbf{44.9} \small{($\pm$2.72)} &\textbf{55.1} \small{($\pm$1.27)}&\textbf{43.8} \small{($\pm$1.22)}\\
         \cline{3-7}
         &66.3/47.0&$\Delta$&+3.6&+1.0&+35.6&+29.6\\
         \hline
         \multirow{10}{*}{\rotatebox[origin=c]{90}{SVHN}}&\textbf{4-layer CNN}&Parameters& \multicolumn{2}{c|}{0.106 M (50\%)} & \multicolumn{2}{c}{0.065 M (30\%)}\\
         \cline{3-7}
         &Size=0.215 M &Hydra&47.8&33.9&19.6&19.6\\
         &Standard/&FaShapley&49.5&37.2&41.9&32.3\\
         &Verified &VeriCompress&\textbf{57.8} \small{($\pm$2.45)}&\textbf{38.8} \small{($\pm$2.34)} &\textbf{54.9} \small{($\pm$3.33)}&\textbf{37.5} \small{($\pm$1.42)}\\
         \cline{3-7}
         &60.0/41.4&$\Delta$&+8.3&+1.6&+13.0&+5.2\\
         \cline{2-7}
         &\textbf{7-layer CNN} &Parameters& \multicolumn{2}{c|}{3.295 M (20\%)} & \multicolumn{2}{c}{2.239 M (15\%)}\\
         \cline{3-7}
         &Size=17.190 M &Hydra&49.2&34.9&15.9&15.9\\
         &Standard/&FaShapley&60.1&43.6&39.0&32.0\\
         &Verified&VeriCompress&\textbf{68.7} \small{($\pm$1.35)}&\textbf{49.4} \small{($\pm$1.94)}&\textbf{66.9} \small{($\pm$4.95)}&\textbf{47.7} \small{($\pm$4.38)}\\
         \cline{3-7}
         &73.8/55.4&$\Delta$&+8.6&+5.8&+27.9&+15.7 \\
         \hline
    \end{tabular}}
    \caption{Standard and Verified Accuracy for 4-layer CNN and 7-layer CNN trained for CIFAR-10 and SVHN at $\varepsilon= 2/255$ for different parameters budgets (corresponding to structured pruning amounts used in the baseline). $\Delta$ represents the change as compared to the best baseline, M stands for million parameters}
    \label{tab:structured-2-255}
    \vskip -2ex
\end{table}

 \subsection{Establishing the Wider Applicability}

This section demonstrates the applicability and scalability of \textcolor{violet}{VeriCompress} by evaluating various \emph{backbone network} architectures, complexities, and datasets. Evaluation results are discussed below:
 
$\bullet$ \textbf{Evaluation for Complex Models:}
\label{larger_models}
To demonstrate the scalability of \textcolor{violet}{VeriCompress} to more complex backbone network architectures, we compute compressed models for Resnet, DenseNet, and ResNext (used by \cite{auto-lirpa}) on the CIFAR-10 dataset. Table \ref{tab:larger-models-structured} shows that \textcolor{violet}{VeriCompress} extracted compressed models’ performance in comparison to their dense counterparts is comparable to the Table \ref{tab:structured-2-255} evaluation results, showing \textcolor{violet}{VeriCompress} performs similarly across different network complexities and architectures.

\begin{table}[t]
    \centering
    \resizebox{0.99\linewidth}{!}{
    \begin{tabular}{c|cc|cc|cc}
        \hline
         Model & Standard & Verified& Standard & Verified&Standard & Verified \\
         \hline
         \textbf{Resnet}&\multicolumn{2}{c|}{4.213 M (100\%)}& \multicolumn{2}{c|}{0.799 M (20\%)} & \multicolumn{2}{c}{0.398 M (10\%)} \\
         \cline{2-7}
         &57.2&46.0&55.0&44.1&51.0&41.5\\
        
         \hline
         \textbf{DenseNet}&\multicolumn{2}{c|}{7.897 M (100\%)}& \multicolumn{2}{c|}{1.667 M (20\%)} & \multicolumn{2}{c}{0.914 M (10\%)}\\
         \cline{2-7}
         &57.7&46.8&56.5&45.3&49.2&38.9\\
         \hline
         \textbf{ResNeXt}&\multicolumn{2}{c|}{17.591 M (100\%)}& \multicolumn{2}{c|}{3.330 M (20\%)} & \multicolumn{2}{c}{1.714 M (10\%)}\\
         \cline{2-7}
         &58.5&46.7&58.2&46.5&45.8&37.2\\
         \hline
    \end{tabular}}
    \caption{Comparison of compressed models obtained using \textcolor{violet}{VeriCompress} for Resnet, DenseNet and ResNeXt trained for CIFAR-10 at  $\varepsilon = 2/255$ with their dense counterparts.}
    \label{tab:larger-models-structured}
\end{table}

$\bullet$ \textbf{Evaluations for more Datasets:}
This section evaluates \textcolor{violet}{VeriCompress}'s generalizability on MNIST and Pedestrian Detection). Table \ref{tab:svhn_pedestrain_detection} demonstrates the results for additional image datasets: (a) a pedestrian detection dataset that aims at differentiating between people and people-like objects \cite{pedestrian-dataset}, and (b) MNIST. The backbone architecture used for these evaluations is a 7-layer CNN, adapted to the respective datasets. For both datasets, the performance of the compressed models is comparable to that of their dense counterparts. The perturbation amount used for Pedestrian Detection and MNIST are  $2/255$, and $0.4$. 

\begin{table}[t]
    \centering
    \resizebox{0.99\linewidth}{!}{
    \begin{tabular}{c|cc|cc|cc}
        \hline
         Dataset & Standard & Verified& Standard & Verified&Standard & Verified \\
         \hline
         \textbf{Pedestrian}&\multicolumn{2}{c|}{321.5 M (100\%)}& \multicolumn{2}{c|}{122.6 (40\%)} & \multicolumn{2}{c}{90.5 M (30\%)}\\
         \cline{2-7}
         \textbf{Detection}&71.6&36.5&71.1&33.6&68.7&24.2\\
         \hline
         \textbf{MNIST}&\multicolumn{2}{c|}{1.97 M (100\%)}& \multicolumn{2}{c|}{0.94 M (50\%)} & \multicolumn{2}{c}{0.74 M (40\%)}\\
         \cline{2-7}
         &97.7&87.8&95.4&82.2&95.1&80.1\\
         \hline
    \end{tabular}}
    \caption{Comparison of compressed models obtained using \textcolor{violet}{VeriCompress} using 7-layer CNN as backbone architecture trained for (a) Pedestrian Detection $(\varepsilon = 2/255)$ and (b) MNIST $(\varepsilon = 0.4)$}
    \label{tab:svhn_pedestrain_detection}
    \vskip -3ex
\end{table}

\subsection{Training and Inference Resource Reduction}\label{Eval:Resource-reduction}
This section evaluates the training and inference time, and resource reduction through the \textcolor{violet}{VeriCompress} extracted compressed NNs, compared to their dense counterparts and baselines.

\begin{table}[t]
    \centering
    \resizebox{0.7\linewidth}{!}{
    \begin{tabular}{c|c|c|c}
        \hline 
        Method $\Rightarrow$&Hydra&FaShapely&VeriCompress\\
        \hline 
        4-layer CNN &9.2&9.0&3.4\\
        7-layer CNN & 20.9&18.8&7.5\\
         \hline
    \end{tabular}}
    \vskip -2ex
    \captionof{table}{Training Time (in Ksec) for Hydra, FaShapley and VeriCompress}
    \label{tab:training-time}
\end{table}

\subsubsection{Reduction in Training Time for \textcolor{violet}{VeriCompress}'s Compressed Network Extraction Compared to the Baselines:} 
\textcolor{violet}{VeriCompress} requires almost one-third of clock-time as that of the pruning-based baselines~\cite{sehwag2020hydra,kang2023fashapley}. Table \ref{tab:training-time} compares the training time required by Hydra~\cite{sehwag2020hydra}, FaShapley~\cite{kang2023fashapley} and \textcolor{violet}{VeriCompress} for 4-layer CNN and 7-layer CNN trained for CIFAR-10, while using the same number of training epochs. That is, the total number of epochs used in pre-training, pruning, and fine-tuning for state-of-the-art \cite{sehwag2020hydra,kang2023fashapley} and an equal number of epochs for \textcolor{violet}{VeriCompress}. These training times are computed for NVIDIA RTX A6000.

\begin{table}[t]   
    \vskip -1ex
    \centering
    \resizebox{0.75\linewidth}{!}{
    \begin{tabular}{c|c|c|c}
        \hline 
        \textbf{Model}&Compression $\Rightarrow$&0\%&90\%\\
        \hline 
        \multirow{4}{*}{7-layer CNN}& Inference Time(ms) &6.4&3.5\\
        &Memory(MB)&68.8&13.7\\
        &Peak CPU usage (\%) & 55 &51\\
        &Peak RAM usage(Mb) & 98&60\\
        \hline
        \multirow{4}{*}{Resnet}& Inference Time(ms) &4.54&0.81\\
        &Memory(MB)&16.9&2.07\\
        &Peak CPU usage(\%) & 54 &51\\
        &Peak RAM usage(Mb) & 50&37\\
         \hline
    \end{tabular}}
    \caption{Comparison of inference time per sample and resource requirements for compressed models for 7-layer CNN and Resnet trained for CIFAR-10 with their dense counterparts on Google Pixel 6}
    \vskip -3ex
    \label{tab:structured_time_and_memory}
\end{table}

\subsubsection{Real-deployment Evaluations on the Reduction in Time and Memory Requirement of Compressed Models: }

\textcolor{violet}{VeriCompress}'s compressed models result in memory and inference time reductions in the resource constraint generic platforms.  Table \ref{tab:structured_time_and_memory} compares compressed models with their dense counterparts on Google Pixel 6, which has 8GB RAM, powered by a 2.8GHz octa-core Google Tensor processor, running on an Android 12.0 system. 
\par 
The inference time shown in Table \ref{tab:structured_time_and_memory} is computed as the average over the result of three experiment trials, which computes the average time required per sample over 10,000 repetitions.  It is observed that compressed 7-layer CNN and Resnet require about $5-8$ times less memory, $2-4$ times less inference time, and significantly less RAM used by their dense counterparts. These evaluations demonstrate the impact of \textcolor{violet}{VeriCompress} in enabling resource-constraint generic platforms to leverage verified robust models. However, the resource reductions come at the cost of reduced generalizability which is discussed in Appendix \ref{cost-benefit-analysis}.

\section{Real World Deployment Scope and Impact}
Medical diagnosis and other safety-critical applications such as autonomous driving need a guarantee of accurate inferences while having a model complying with memory constraint \cite{safety-critical}. 
However, training such compressed NN models using state-of-the-art structured pruning approaches requires in-depth knowledge of pruning and extensive computation resources. In addition, tuning a large number of hyper-parameters during training and network structure identification increases the computation time exponentially. This complexity might not be practical for end-users, including developers in domains like internet-of-things (IoT), embedded systems, or medical applications.  
Following \textcolor{violet}{VeriCompress's} traits make it a viable solution for such scenarios :

\par $\bullet$ Automates the learning of compressed models: \textcolor{violet}{VeriSprase} enables the user to give as input a complex model architecture and a parameter budget that depends on the memory constraint of the deployment platform and generates a compressed model which exhibits high verified robustness. Thus, the user doesn't need the expertise of model pruning mechanisms to achieve the compressed model.


\par $\bullet$ Faster training: Notably, the training time of the \textcolor{violet}{VeriCompress} approach is $2-3$ times less than the baseline structured pruning approaches. Furthermore, \textcolor{violet}{VeriCompress} computes the importance of a model element (kernel/node) using the magnitude of its parameters, which is data independent; thus, requires significantly less computation as compared to state-of-the-art approaches that rely on specific data-dependent importance scores training phases. 

\par $\bullet$ Fewer hyper-parameters to tune: In addition to common hyper-parameters such as batch size, learning rate, etc., verified robustness approaches generally require hyper-parameters such as the number of epochs and start-and-end of perturbation scheduler. State-of-the-art structured pruning approaches Hydra and Fashapley need to tune these hyperparameters for 3 phases: pre-training, pruning, and finetuning. In contrast, \textcolor{violet}{VeriCompress} reduces that to a single phase.

\section{Conclusion}
This paper presents \textcolor{violet}{VeriCompress} - a tool designed to streamline the synthesis of compressed verified robust NNs from scratch, leveraging relatively less computation, time, and effort than the state-of-the-art literature. Our empirical evaluations demonstrate the effectiveness of \textcolor{violet}{VeriCompress} in achieving compressed networks of the required size while exhibiting generalizability comparable to its dense counterparts. The practical deployment of these resultant models onto a resource-constraint generic platform effectively demonstrates reductions in both memory and computational demands. The accelerated generation process, broader applicability, and effectiveness of \textcolor{violet}{VeriCompress} will greatly assist users such as edge, embedding systems, or IoT developers in effortlessly generating and deploying verified robust compressed models tailored for their respective safety-critical and resource-constrained generic platforms.

\section*{ACKNOWLEDGMENTS}
This work was partly supported by NSF IIS SCH \#2124285 and NSF CNS CPS \#2148187.

\bibliography{aaai24}

\newpage
\appendix

\section{A. Background and Related-Works}
\label{Background-and-motivation}
\par
\subsection{Background on Formal robustness verification} 

\emph{An NP-Complete problem:} The complexity and the need for non-linear activation functions in recent NNs to learn complex decision boundaries makes the problem of formal robustness verification of NNs non-convex and NP-Complete ~\cite{reluplex}.

\emph{Relaxation-based Methods:} To address the challenge, \citet{reluplex} used relaxation of ReLU activations, temporarily violating the ReLU semantics, resulting in an efficient query solver for NNs with ReLU activations. The piece-wise linearity of the ReLU activations is the basis for this relaxation. Linear Relaxation-based Perturbation Analysis (LiRPA) based methods compute the linear relaxation of a model containing general activation functions which are not necessarily piece-wise linear and verify non-linear NNs in polynomial time. Thus, LiRPA-based methods use Forward (IBP~\cite{Gowal2018OnTE}), Backward (CROWN~\cite{crown}, Fastened-CROWN~\cite{Lyu2020FastenedCT}) or Hybrid (CROWN-IBP~\cite{crown-ibp}) bounding mechanisms to compute linear relaxation of a model. CROWN~\cite{crown} uses a backward bounding mechanism to compute linear or quadratic relaxation of the activation functions. IBP~\cite{Gowal2018OnTE} is a forward bounding mechanism that propagates the perturbations induced to the input towards the output layer in a forward pass based on interval arithmetic, resulting in ranges of values for each class in the output layer. CROWN-IBP~\cite{crown-ibp} is a hybrid approach that demonstrated that IBP can produce loose bounds, and to achieve tighter bounds, it uses IBP ~\cite{Gowal2018OnTE} in the forward bounding pass and CROWN~\cite{crown} in the backward bounding pass to compute bounding hyperplanes for the model. CROWN-IBP has been shown to compute tighter relaxations, thus, providing better results for formal verification. 

Specifically, for a sample $x_0$, the perturbation neighborhood is defined as $\ell_p$-ball of radius $\varepsilon$~\cite{crown}: 
\begin{equation}
    \label{lp-ball}
\mathbb{B}_p(x_0,\varepsilon) := \{x | \| x - x_0\|_p \leq \varepsilon\}.
\end{equation}
LiRPA aims to compute linear approximations $f^\varepsilon$ of the model $\cM_\theta$ ($\theta$ represents model parameters), and provide lower $f_L^\varepsilon$ and upper $f_U^\varepsilon$ bounds at the output layer, such that for any sample $x \in \mathbb{B}_p(x_0,\varepsilon)$ and each class $j$, the model output is guaranteed to follow:
\begin{equation}
    f_L^\varepsilon(x)^j \leq \cM_\theta(x)^j \leq f_U^\varepsilon(x)^j.
\end{equation} 

\subsection{Background on Verified robustness.}\label{VLR-def}
A NN model, $\cM_\theta$ is said to be verified robust ~\cite{fromherz2021fast} for a sample $(x_0,y)$ if: 
$\forall x \in \mathbb{B}_p(x_0,\varepsilon) \Rightarrow \cM_\theta(x) = \cM_\theta(x_0)$, that is, for all the samples $x \in \mathbb{B}_p(x_0,\varepsilon)$, the model is guaranteed to assign the same class as it assigns to $x_0$. LiRPA approaches measure the    robustness in terms of a margin vector $m(x_0,\varepsilon) = Cf^\varepsilon(x_0)$, where C is a \emph{specification matrix} of size $n\times n$ and $n$ is the number of all possible classes~\cite{Gowal2018OnTE,crown,crown-ibp,auto-lirpa}. \citet{crown-ibp} defines $C$ for each sample $(x_0,y)$ as:
\begin{equation}
    C_{i,j} = 
    \begin{cases}
    1, & j = y \text{ and } i \ne y \text{ (truth class)}\\
    -1, & i = j \text{ and } i \ne y \text{ (other classes)}\\
    0, &\text{ otherwise}.
    \end{cases}
\end{equation}

The entries in matrix $C$ depend on the true class $y$. In matrix $C$, the row corresponding to the true class contains $0$ at all the indices. All the other rows contain $1$ at the index corresponding to the true class $(j = y)$ and $-1$ at the index corresponding to current class $(j = i)$ and $0$ at all the other indices. Thus, the $i^{th}$ value of $m(x_0,\varepsilon)$ is given by $m^i(x_0,\varepsilon) = f^\varepsilon(x_0)^y - f^\varepsilon(x_0)^i$, which is the difference of output values of the true class $y$ from the output value corresponding to class $i$.

Further, $\underline{m}(x_0,\varepsilon)$ represents the lower bound of the margin vector. If all the values of the $\underline{m}(x_0,\varepsilon)$ are positive, $\forall_{i\ne y}\underline{m}^i(x_0,\varepsilon) > 0$, the model $\cM_\theta$ is said to be   robust for sample $x_0$.  That implies, the model will always assign the highest output value to the true class label $y$ if a perturbation less than or equal to $\varepsilon$ is induced to the sample $x_0$. Thus, $\forall_j \underline{m}(x_0,\varepsilon)^j > 0$
implies the guaranteed absence of any adversarial sample in the region of interest. Furthermore, to specify the region bounded by $\mathbb{B}_p(x_0,\varepsilon)$ , we use $p =\infty$ because $\ell_\infty$-norm covers the largest region around a sample and is most challenging and widely-used~\cite{kang2023fashapley,wongNKolter}. 

\subsection{Background on Model training to maximize robustness}
\label{  -robustness}
Since a model is considered verified robust for a sample $(x_0,y)$ if all the values of $\underline{m}(x_0,\varepsilon) > 0$, the training schemes proposed by previous approaches~\cite{Gowal2018OnTE,crown-ibp} aim to maximize robustness of a model by maximizing the lower bound $\underline{m}(x_0,\varepsilon)$. \citet{auto-lirpa}  defines the training objective as minimizing the maximum cross-entropy loss  between the model output $\cM_\theta(x)$ and the true label $y$ among all samples $x \in  \mathbb{B}_p(x_0,\varepsilon)$ (eq. 6 in ~\cite{auto-lirpa}). Thus, the training loss is defined as:
\begin{equation}
    \cL_\text{train}(\cM_\theta,\cD,\varepsilon) = \sum_{(x_0,y) \in \cD} \max_{x\in\mathbb{B}(x_0,\varepsilon)}\cL(\cM_\theta(x),y),
    \label{eq:robust_loss}
\end{equation}
where $\cD$ is the dataset and $(x_0,y) \in \cD$. Intuitively, $\cL_\text{train}$ is the sum of maximum cross-entry loss in the neighborhood of each sample in $\cD$ for perturbation amount $\varepsilon$.
\citet{wongNKolter} showed that the problem of  minimizing $\cL_\text{train}$ (as defined in \ref{eq:robust_loss}) and the problem of maximizing $\underline{m}(x_0,\varepsilon)$ are dual of each other. Thus, \emph{a solution that minimizes $\cL_\text{train}$, maximizes the values of $\underline{m}(x_0,\varepsilon)$, hence maximizes the    robustness for the sample $(x_0,y)$}. Since, \emph{$x_0 \in \mathbb{B}_p(x_0,\varepsilon)$, minimizing $\cL_\text{train}$ also maximizes benign/standard accuracy, ergo maximizes \emph{generalizability} altogether}.

\citet{auto-lirpa} computes $\max_{x\in\mathbb{B}(x_0,\varepsilon)}\cL(\cM_\theta(x),y)$ which requires computing bounding hyper-planes $f_U^\varepsilon$ and $f_L^\varepsilon$ using one of the aforementioned LiRPA bounding techniques. 
Following \citet{auto-lirpa}, \textcolor{violet}{VeriCompress} employs CROWN-IBP hybrid approach as the bounding mechanism to optimize $\cL_\text{train}(\cM_\theta,\cD,\varepsilon)$ in equation \eqref{eq:robust_loss}. The perturbation amount $\varepsilon$ is initially set to zero and gradually increases to $\varepsilon_\text{max}$ according to the perturbation scheduler $\varepsilon$-scheduler$(\varepsilon_\text{max},t,s,l)$ (Section \ref{presented-approach}).

\paragraph{$\bullet$ Perturbation Scheduler $(\varepsilon$-scheduler ($\varepsilon_\text{max},t,s,l)$:}
\label{appendix:perturbation_scheduler}
$\varepsilon$-scheduler provides a perturbation amount for every training epoch $t\geq s$, which gradually increases $\varepsilon$ starting at epoch s. The schedule starts with a 0 perturbation and reaches $\varepsilon_\text{max}$ in $l$ epochs~\cite{Gowal2018OnTE}. According to the literature \cite{Gowal2018OnTE,crown-ibp,auto-lirpa}, the Linear Relaxation-based Perturbation Anaysis (LiRPA) approaches use interval propagation arithmetic to propagate input perturbation to the output layer, where the interval size usually keeps increasing as the propagations reach deeper in the model. This issue is known as the problem of intermediate-bound explosion. A perturbation scheduler helps deal with this issue by gradually increasing the perturbation amount.

\section{Model Compression} 
While various compression techniques are available, to the best of our knowledge, only model pruning approaches have been leveraged to achieve verified robust compressed models in the literature ~\cite{sehwag2020hydra,kang2023fashapley}. Consequently, the focus of this section is confined to pruning discussions. Furthermore, given that \textcolor{violet}{VeriCompress} adopts the Dynamic Sparse Training (DST) approach, a discussion on DST is also included.

\subsection{Model Pruning} Model pruning approaches train a dense model to convergence and remove (or prune) the parameters, either in one step or iteratively,  which contribute the least towards model inference. To compensate for the information lost during pruning, an additional step of fine-tuning the retained parameters is performed by associating a static binary mask with the parameters. The removal of parameters can be done at different granularity: unstructured pruning (fine-grained) and structured pruning (coarse-grained).

\subsection{Structured vs. Unstructured Pruning} Unstructured pruning zeros out the parameters depending on their individual importance~\cite{sehwag2020hydra}, whereas structured pruning removes channels/nodes as a whole~\cite{cstar}. It has been observed that structured pruning is effective in significantly reducing memory and inference time in generic platforms, making them highly effective in practical use \cite{no-time-reduction}.

\subsubsection{Selecting the Parameters to be Removed during Sparsification:}
\label{appendix:param_select} 
Selection of parameters to be removed can be done based on several criteria such as the magnitude of weights~\cite{sehwag2019compact}, the norm of corresponding gradients~\cite{grad_prune}, etc. Additionally, recent approaches associate importance scores with each parameter based on the respective training objective~\cite{sehwag2020hydra,kang2023fashapley}, and remove the parameters with the least important scores. Presented \textcolor{violet}{VeriCompress} exceeds state-of-the-art without any task/objective specific important score-based parameter selection mechanism, establishing that a compressed network with high verified local robustness and accuracy comparable to its dense counterpart is attainable from scratch following trivial parameter selection mechanism.

\subsection{Dynamic Sparse Training} Recently developed \emph{dynamic-mask-based sparse-training} known as Dynamic Sparse Training (DST)~\cite{nest,itop} approaches follow \emph{grow-and-prune} paradigm to achieve a favorable sparse network starting from a random sparse \emph{seed network}. \citet{nest} grows the model to a certain capacity and then gradually removes the connection, following a single \emph{grow-and-prune} stride. However, recent approaches~\cite{itop} employ multiple repetitions of \emph{grow-and-prune} where, in each repetition, new connections (i.e., parameters) which minimize the natural loss are added, and the least important connections are removed.


\section{Code}

The code is available at https://github.com/Sawinder-Kaur/VeriCompress.

\section{Cost-Benerfit Analysis}
\label{cost-benefit-analysis}
It is evident from Table 1 (Section 4) that while the presented \textcolor{violet}{VeriCompress} approach surpasses existing methods, reducing the number of parameters still negatively impacts model performance, which is measured in terms of accuracy and verified robustness. For instance, for 7-layer CNN trained for CIFAR10, when the number of parameters is reduced from 17.19 M to 3.26M (approximately 5 times reduction than the over-parameterized model), the accuracy and verified robustness reduces by 6.4\% and 2.1\% respectively. However, a reduction to 1.58M (approximately 10 times reduction than the over-parameterized model) results in a drop of 11.2\% in accuracy and 3.2\% in verified robustness. Additionally, this drop varies with the complexity and the architecture of the backbone model. Notably, as shown in Table 4 (Section 4) of the main paper, fewer parameters lead to smaller-sized models and reduced inference time. Thus resulting in efficient models. For instance, reducing the parameters of the 7-layer to one-tenth reduces the memory requirement to one-fifth and inference time to half.

\begin{figure}[t]
    \centering
    \subfigure[4-layer CNN (CIFAR-10)]{\includegraphics[width=0.50\linewidth]{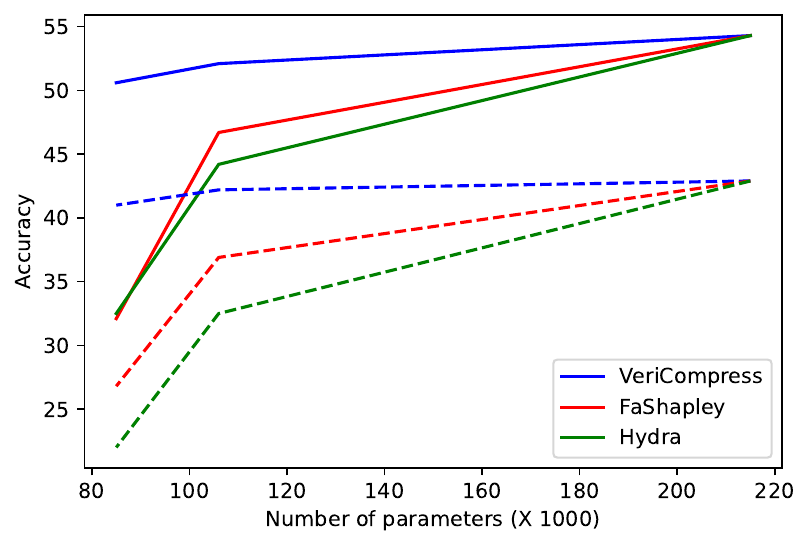}}\subfigure[7-layer CNN (CIFAR-10)]{\includegraphics[width=0.45\linewidth]{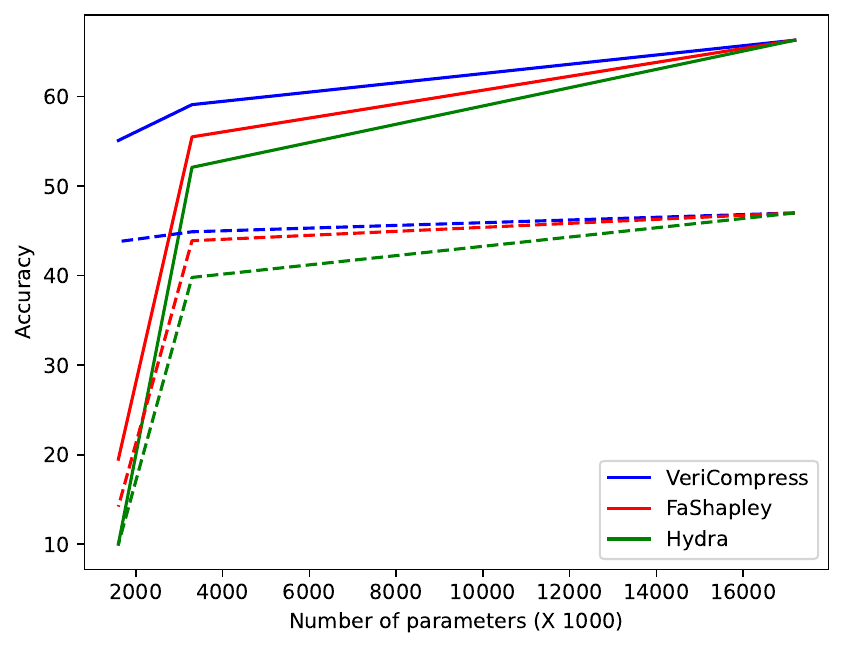}} 
    \subfigure[4-layer CNN (SVHN)]{\includegraphics[width=0.48\linewidth]{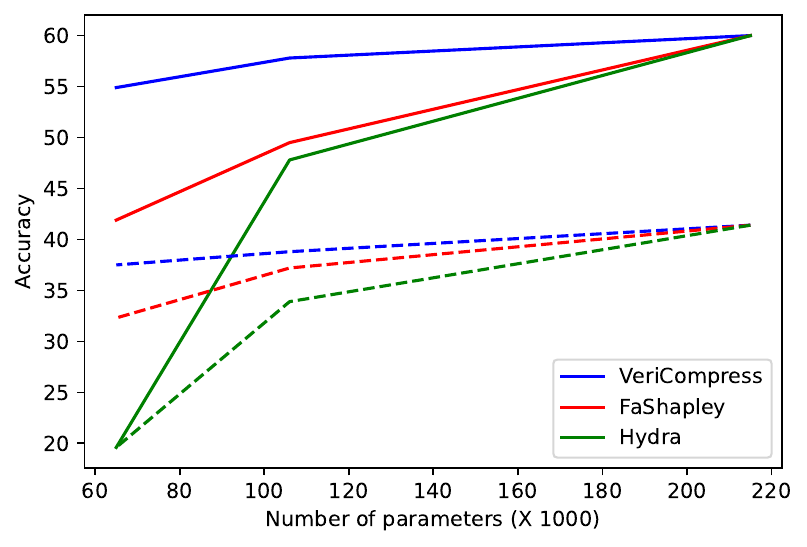}}
    \subfigure[7-layer CNN (SVHN)]{\includegraphics[width=0.48\linewidth]{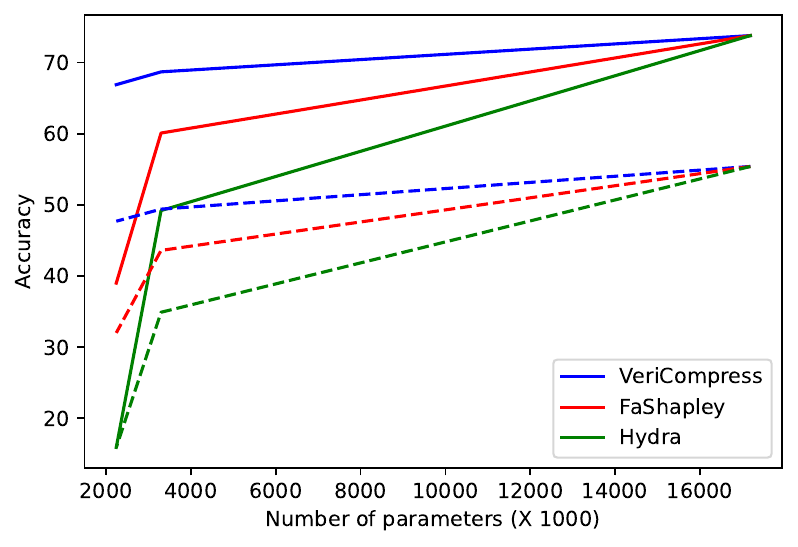}}
    \caption{Cost Benefit Analysis for (a) 4-layer CNN (CIFAR10)  (b) 4-layer CNN (CIFAR-10) (b) 4-layer CNN (SVHN) (d) 7-layer CNN (SVHN). The pertubation amount used for robust training is 2/255.}
    \vskip -2ex \label{fig:cost-benefit}
\end{figure}

Figure \ref{fig:cost-benefit} showcases the superior performance of the VeriCompress approach over the baselines. The figure discusses the cost-benefit analysis for a 4-layer CNN and 7-layer CNN trained for CIFAR-10 and SVHN. The solid and dotted lines represent the accuracy and verified robustness of the model. It's evident that while the presented \textcolor{violet}{VeriCompress} approach surpasses existing methods, reducing the number of parameters still negatively affects model performance, as measured by accuracy and verified robustness. 

This underscores the existence of a trade-off between model efficiency and performance. Consequently, the choice between prioritizing high performance or enhanced efficiency depends on the specific application requirements and the available resources on the deployment platform.

\end{document}